\documentclass[letterpaper]{article} 
\usepackage{aaai25}  
\usepackage{times}  
\usepackage{helvet}  
\usepackage{courier}  
\usepackage[hyphens]{url}  
\usepackage{graphicx} 
\usepackage{natbib}  
\usepackage{caption} 
\usepackage{multirow}
\usepackage{booktabs}
\usepackage{algorithm}
\usepackage{algorithmic}

\usepackage{newfloat}
\usepackage{listings}
\DeclareCaptionStyle{ruled}{labelfont=normalfont,labelsep=colon,strut=off} 
\floatstyle{ruled}
\newfloat{listing}{tb}{lst}{}
\floatname{listing}{Listing}
\title{Adaptive PII Mitigation Framework for Large Language Models}

\author{
    Shubhi Asthana, 
    Ruchi Mahindru, 
    Bing Zhang,
    Jorge Sanz
}
\affiliations{
    IBM Research\\

    sasthan@us.ibm.com,
    rmahindr@us.ibm.com,
    jorges@us.ibm.com,
    bing.zhang@ibm.com
}

\usepackage{bibentry}

\begin{document}

\maketitle

\begin{abstract}

Artificial Intelligence (AI) faces growing challenges from evolving data protection laws and enforcement practices worldwide. Regulations like GDPR and CCPA impose strict compliance requirements on Machine Learning (ML) models, especially concerning personal data use. These laws grant individuals rights such as data correction and deletion, complicating the training and deployment of Large Language Models (LLMs) that rely on extensive datasets. Public data availability does not guarantee its lawful use for ML, amplifying these challenges.

This paper introduces an adaptive system for mitigating risk of Personally Identifiable Information (PII) and Sensitive Personal Information (SPI) in LLMs. It dynamically aligns with diverse regulatory frameworks and integrates seamlessly into Governance, Risk, and Compliance (GRC) systems. The system uses advanced NLP techniques, context-aware analysis, and policy-driven masking to ensure regulatory compliance.

Benchmarks highlight the system’s effectiveness, with an F1 score of \textit{0.95} for \textit{Passport Numbers}, outperforming tools like Microsoft Presidio \textit{(0.33)} and Amazon Comprehend \textit{(0.54)}. In human evaluations, the system achieved an average user trust score of \textit{4.6/5}, with participants acknowledging its accuracy and transparency. Observations demonstrate stricter anonymization under GDPR compared to CCPA, which permits pseudonymization and user opt-outs. These results validate the system as a scalable and robust solution for enterprise privacy compliance.

\end{abstract}

%

\section{Introduction}
\label{sec:Introduction}
The rapid advancement of Artificial Intelligence (AI) has revolutionized industries but also introduced significant challenges in ensuring compliance with evolving global data protection regulations. Large Language Models (LLMs), which rely on vast amounts of publicly sourced data, are particularly susceptible to legal complexities surrounding the use of personal data \cite{chen2023can}. Regulatory frameworks such as the General Data Protection Regulation (GDPR)\cite{gdpr}, the California Consumer Privacy Act (CCPA) \cite{ccpa}, and Canada's Personal Information Protection and Electronic Documents Act (PIPEDA) \cite{pipeda} demand stringent safeguards to ensure data privacy, often imposing jurisdiction-specific requirements. The GDPR enforces strict data protection rules across the EU and applies extraterritorially, impacting organizations worldwide that process the personal data of EU residents. For example, a U.S.-based e-commerce site must comply with GDPR if it offers goods or services to EU customers, emphasizing principles like data minimization and explicit user consent. The CCPA on the other hand, prioritizes consumer rights, granting California residents control over their data through rights such as accessing personal data, requesting its deletion, or opting out of data sales. For instance, a social media platform serving California users must allow them to opt out of targeted advertising based on their data. PIPEDA in Canada focuses on accountability and meaningful consent. For example, a Canadian bank collecting customer information for a loan application must transparently explain why the data is being collected, how it will be used, and ensure secure handling, especially during business transactions like mergers.

The use of publicly available data in LLMs like online publications, social media, web-crawled data etc. often carries implicit risks, as public availability does not equate to legal authorization for secondary use in AI training model \cite{yanamala2024navigating}. In jurisdictions such as the European Union and Canada, explicit consent is required for new uses of publicly available personal data, even when the data is non-private. Conversely, some U.S. states permit the use of such data without additional consent but uphold rights like the \textbf{right-to-be-forgotten} and \textbf{right-to-amend}, as emphasized under the CCPA \cite{privacy}. 

Global enterprises must adopt strong risk mitigation strategies to navigate these legal complexities, as AI models using internet-sourced data must comply with varying regulations based on data provenance and deployment location. LLMs, in particular, require advanced technologies to avoid compromising personal data \cite{oluokun2024building}.

These disparities pose a critical challenge for global enterprises deploying AI systems across multiple jurisdictions. Failure to adhere to these complex regulatory landscapes can result in legal repercussions, reputational harm, and compromised user trust. Therefore, mitigating the risks associated with Personally Identifiable Information (PII) and Sensitive Personal Information (SPI) is paramount.

This paper proposes an adaptive system designed to dynamically assess and mitigate risks associated with PII and SPI in LLMs. By incorporating advanced \textit{Natural Language Processing (NLP), real-time contextual analysis}, and \textit{policy-driven remediation mechanisms}, the system ensures compliance with diverse regulations while aligning with enterprise Governance, Risk, and Compliance (GRC) policies. The framework also addresses industry-specific needs:

\begin{enumerate}
    \item \textbf{Healthcare}: Safeguards patient data in AI-driven healthcare solutions, ensuring compliance with Health Insurance Portability and Accountability Act (HIPAA) \cite{riad2024enhancing}.
    \item \textbf{Finance}: Protects sensitive financial data, aligning with regulations such as GDPR, CCPA, and the Banking Secrecy Act (BSA) \cite{yanamala2023evaluating}.
\end{enumerate}


While offering a scalable approach to mitigating regulatory risks, the system must address the following challenges: 
\begin{enumerate}
\item continuous evolution of data protection laws
\item integration of conflicting jurisdictional requirements
\item accurate identification of PII and SPI across languages and cultural contexts.
\end{enumerate}

This paper contributes to advancing ethical AI deployment by presenting a robust solution that meets current regulatory demands while anticipating future challenges. The major contributions of this work are:

\begin{itemize}
    \item Adaptive Risk Mitigation Framework: Introduces a novel system to dynamically identify and mitigate risks associated with PII and SPI in LLMs, ensuring compliance with global data privacy regulations.
    \item Real-Time Compliance Mechanisms: Develops real-time contextual analysis and policy-driven remediation to align AI systems with jurisdiction-specific laws like GDPR, CCPA, and PIPEDA.
    \item Industry-Specific Applications: Demonstrates the framework’s effectiveness in safeguarding sensitive data in critical sectors like healthcare (HIPAA compliance) and finance (BSA and GDPR compliance).
    \item Enterprise Alignment: Integrates with Governance, Risk, and Compliance (GRC) policies, offering a scalable solution for global AI deployments.
\end{itemize}

\paragraph{Motivation for an Adaptive Framework}
Existing solutions for PII protection, such as static rule-based systems or simple redaction tools, are inadequate in addressing the dynamic nature of privacy regulations and the complexity of modern datasets. There is a pressing need for: 
\begin{itemize}
    \item \textbf{Dynamic Regulatory Automation:} Adaptive systems that dynamically integrate and enforce jurisdiction-specific rules.
    \item \textbf{Intelligent Redaction:} Context-Aware mechanisms to ensure accurate and relevant remediation of PII and SPI. 
    \item \textbf{Domain-Agnostic Scalability:} Scalability across use cases to accommodate the diverse applications of LLMs in legal, healthcare, finance, and other domains.
\end{itemize}

This paper aims to fill these gaps by introducing a system that leverages advanced NLP techniques and a regulatory policy engine to address privacy risks proactively. The proposed approach ensures compliance, minimizes regulatory liabilities, and upholds ethical standards in the deployment of AI technologies.

\section{Background and Motivation}
\label{sec:Background}


The rise of AI technologies has coincided with a rapid increase in data collection and use. LLMs like GPT-4 and Bard process and generate human-like text by analyzing massive datasets sourced from the internet. While this has unlocked transformative possibilities across many fields, it has also amplified existing privacy and compliance challenges. These challenges aren’t new—personal data has long been publicly available. However, the way LLMs reason over this data and extract insights has brought these issues to the forefront, exposing them in ways that were previously less apparent or manageable.

\paragraph{Legal and Ethical Implications}
Data privacy laws such as GDPR and CCPA mandate that organizations processing personal data adhere to principles of lawfulness, fairness, and transparency, requiring explicit consent for data usage \cite{lescrauwaet2022adaptive}. Additionally, users must retain control over their data, including rights to access, delete, or rectify information. The diversity in regulatory landscapes as well as conflicting requirements creates significant obstacles:


\begin{itemize}
    \item \textbf{Jurisdictional Conflicts:} The GDPR requires explicit consent to use publicly available personal data, while the CCPA allows data use unless the consumer opts out. Balancing these conflicting rules can be challenging.
    \item \textbf{Sector-Specific Regulations:} Industries like healthcare and finance impose stricter data protection requirements, such as HIPAA and BSA, which demand additional safeguards for sensitive data \cite{chakraborty2024global}.
\end{itemize}

\paragraph{Technical Challenges}
Ensuring suitable risk assessments within AI systems like LLMs poses unique technical challenges:
\begin{itemize}
    \item \textbf{Contextual Identification of PII:} Traditional PII detection systems often fail to account for context, such as distinguishing public figures whose names may not warrant masking.
    \item \textbf{Dynamic Policy Integration:} New Regulatory requirements also take place, such as a new personal data law in a state or country. In addition, compliance teams in enterprises often need to adjust some rules. These conditions call for adaptable systems that can incorporate new rules without manual intervention.
    \item \textbf{Multilingual and Cultural Variance:} LLMs trained on global datasets must accurately detect and remediate PII in multiple languages and cultural settings \cite{bohlin2024detection}.
\end{itemize}

\section{Proposed Method}
\label{sec:methodology}

\begin{figure}[]
 \centerline{\includegraphics[scale=0.3]{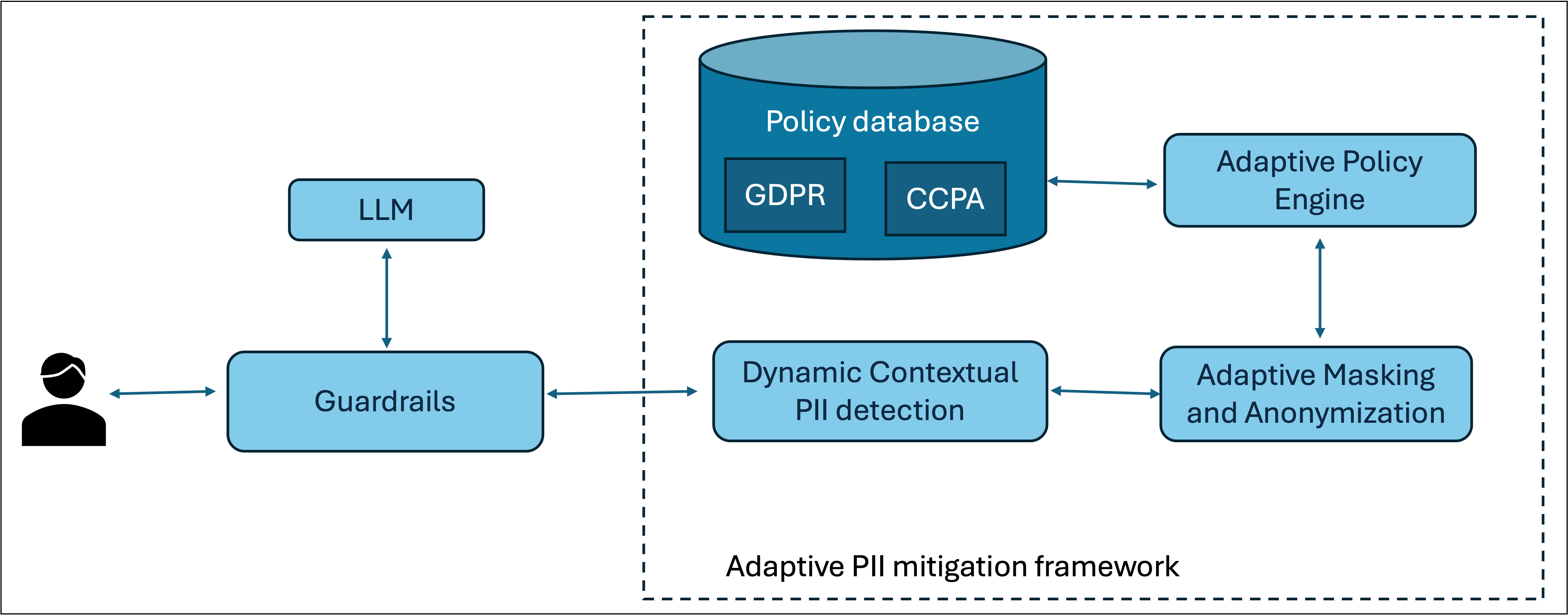}}
\caption{Adaptive PII mitigation framework}
\label{fig:method}
\end{figure}

\begin{table*}[htb]
\footnotesize
    \centering
\begin{tabular}{p{0.1\textwidth} p{0.35\textwidth} p{0.25\textwidth}} 
    \toprule
    \textbf{Level} & \textbf{Description} & \textbf{Example} \\
    \midrule
    Level 1 & Highly sensitive entities (e.g., passport numbers, social security numbers, government IDs). & Passport: E254658340 \\
    Level 2 & Mid-sensitive entities involving personal details linked to a name (e.g., name + date of birth). & Mark Smith’s birthday is July 1st \\
    Level 3 & Less sensitive entities when found without direct linkage to an individual. & Location: San Francisco, Date of Birth: July 1 \\
    \bottomrule
\end{tabular}
\caption{Examples of sensitivity levels}
\label{tab:sensitivity_examples}
\end{table*}

The proposed system leverages an adaptive framework designed to detect and remediate PII and SPI in text data while aligning with global regulatory requirements. The inputs to the system include a Policy Database, which maintains an up-to-date repository of global privacy regulations (e.g., GDPR \cite{gdpr}, CCPA \cite{ccpa}, etc.) The system consists of three key components as seen in Figure \ref{fig:method}:
\begin{enumerate}
    \item \textbf{Adaptive Policy Engine}: Converts regulatory requirements into actionable rules.
    \item \textbf{Dynamic Contextual PII Detection}: Detects PII/SPI with contextual sensitivity analysis.
    \item \textbf{Adaptive Masking and Anonymization Techniques}: Applies regulation-specific remediation strategies to ensure compliance.
\end{enumerate}

The output of the system is the recommended PII/SPI protection strategy for the provided text, tailored to Governance, Risk, and Compliance (GRC) policies.

\subsection{Adaptive Policy Engine}
\textit{Adaptive Policy Engine} dynamically maintains and updates a database of global privacy regulations. Unlike traditional systems that rely on static, hardcoded rules, this engine introduces automation support to help domain experts encode new legal requirements efficiently. This is done by reading documents of legal requirements of policies on a weekly/monthly basis and using a rules-based, machine readable engine to transform it into actionable rules with the participation of domain experts. The system dynamically updates to laws like GDPR, CCPA ensuring that the system remain compliant. The domain expert here adjusts remediation strategies based on conflicting or overlapping requirements across regions (e.g., stricter anonymization for GDPR vs. opt-out mechanisms in CCPA). The set of actionable rules are then evaluated over incoming data into PII detection in second step. 

\subsection{Dynamic Contextual PII Detection}

\begin{figure*}[]
 \centerline{\includegraphics[scale=0.8]{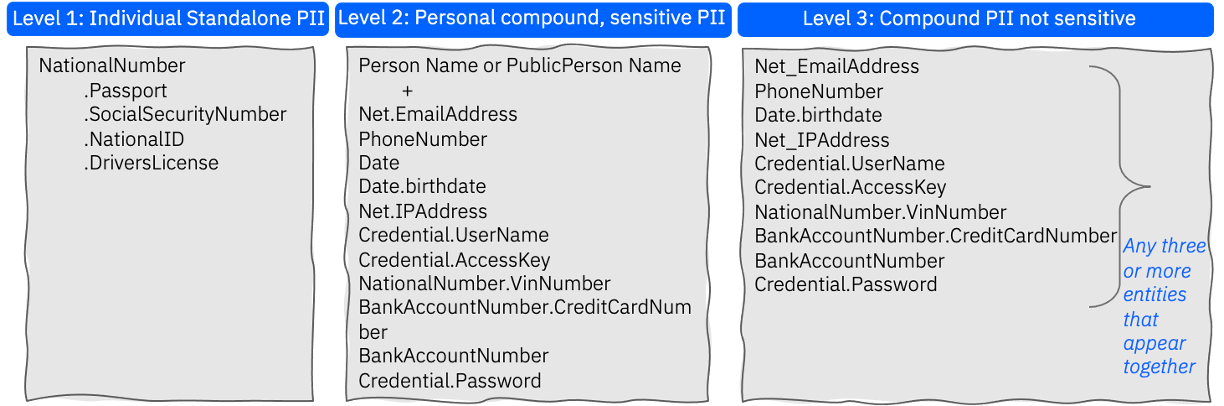}}
\caption{Levels of PII sensitivity}
\label{fig:level}
\end{figure*}

Our novel approach to PII detection involves a multi-step, context-aware process that ensures accurate identification of sensitive data:
\begin{enumerate}
    \item \textbf{Entity Recognition: }Machine learning models identify entities that could potentially be classified as PII or SPI within the text (e.g., names, dates, IDs).
    \item \textbf{Semantic Context Analysis:}
    \begin{itemize}
        \item The system performs semantic analysis to interpret the surrounding context of each detected entity.
        \item For example, the word \textit{"Smith"} in \textit{"John Smith is attending a meeting"} would be flagged as PII, but in "Smith's Bakery offers discounts," it would not.
        \item This step ensures that only sensitive instances of entities are marked, reducing false positives and negatives.
    \end{itemize}
    \item \textbf{Decision-Making for Detection:} The PII detector integrates metadata like usage purpose or geographical location (e.g., GDPR exemptions for public figures) to determine whether an entity should be flagged, masked, or retained.
\end{enumerate}

This nuanced process ensures precision in detecting sensitive information, even in complex or ambiguous contexts.

\paragraph{Contextual Sensitivity Scoring}
Not all detected entities are inherently sensitive. For example:
\begin{itemize}
    \item Example 1: "META celebrates its birthday on July 1st" is non-sensitive as Meta is a public organization.
    \item Example 2: "Mark Smith celebrates his birthday on July 1st" is sensitive as it reveals personal information.
\end{itemize}

To address this, entities are classified into three sensitivity levels based on their context, as shown in Figure \ref{fig:level}. Table \ref{tab:sensitivity_examples} provides examples for these sensitivity levels.

Our algorithm employs proximity analysis to evaluate entities within a defined token range, identifying contextual relationships that influence sensitivity. Detected entities are scored against predefined thresholds corresponding to their sensitivity level, enabling precise and reliable filtering. This scoring mechanism reduces false positives and negatives, outperforming static detection systems by adapting dynamically to the nuances of each case.

Occasionally, an entity may belong to multiple categories, requiring further disambiguation. For example, public figures' names may be exempt from anonymization under GDPR if processed for legitimate interests, but the same names might require masking in contexts involving sensitive disclosures. The algorithm resolves such conflicts by incorporating metadata, such as the purpose of data use or the jurisdictional regulations, to determine the appropriate action dynamically. This context-aware scoring ensures accurate classification and regulatory compliance across a variety of use cases.

\subsection{Adaptive Masking and Anonymization Techniques}

In this step, The adaptive system applies tailored remediation techniques to manage detected PII based on its sensitivity and regulatory requirements. This step ensures that sensitive data is appropriately anonymized or masked while maintaining the usability of the processed data. The primary purpose is to balance regulatory compliance with the need to preserve the integrity and utility of datasets, particularly for downstream applications like machine learning or analytics.

Traditional solutions often rely on generic masking methods such as blanket data removal or static redaction. These approaches face two significant issues:
\begin{enumerate}
    \item Overgeneralization: They treat all PII equally without considering sensitivity or context, leading to either excessive redaction or insufficient protection.
    \item Model Performance Degradation: Excessive data removal or masking can compromise the quality of machine learning models by eliminating context-critical information necessary for accurate predictions or analyses.
\end{enumerate}

To address these challenges, the system dynamically selects remediation techniques based on the specific type of PII and its context. For instance, names are replaced with pseudonyms, allowing readability to be retained while ensuring anonymity. Identifiers, such as Social Security Numbers or passport numbers, are hashed to preserve their uniqueness without exposing sensitive details. Similarly, dates are obfuscated to predetermined year ranges, reducing their granularity to protect privacy while still preserving time-related trends essential for analytical purposes. This context-aware approach ensures that the data remains both compliant and useful.

The system also supports customizable remediation rules, enabling organizations to define and manage compliance strategies tailored to their needs. These rules are not static and require ongoing maintenance to reflect evolving regulations, organizational policies, and operational requirements. To assist users in defining and maintaining these rules:
\begin{enumerate}
    \item Guided Setup: The system provides templates and recommendations based on common regulatory and industry scenarios.
    \item Scalability: It can manage hundreds of rules by categorizing them into regulatory frameworks or operational contexts, ensuring efficient execution and monitoring.
    \item Continuous Updates: As regulations change or business needs evolve, the system incorporates updates to existing rules through an intuitive interface, allowing users to make adjustments without requiring technical expertise.
\end{enumerate}

To ensure the effectiveness and accuracy of these custom rules, the system incorporates a feedback loop. Annotators validate the detection and remediation results, refining the underlying ML models and rule sets. Additionally, the system maintains audit logs of all actions taken on detected PII, offering transparency and accountability. These logs are crucial for regulatory audits, providing a detailed trail of compliance actions and allowing organizations to demonstrate adherence to privacy standards.

By combining adaptability, user-driven customization, and continuous feedback, this framework offers a scalable and robust solution for privacy compliance in dynamic regulatory environments.
\section{Application and Use Cases}
\label{sec:experiments}

The adaptive PII mitigation framework has been applied in various scenarios to address the challenges of personal data protection in LLM-based systems. These applications demonstrate the flexibility and scalability of the framework in production environments.

In practice, there are three scenarios where personal data protection is critical for LLM compliance (regulatory and ethics). \paragraph{LLM Training Data}: The first scenario is about the application of adaptive PII mitigation framework to input data customarily used to train LLMs. While most data sources have public provenance from the Internet, the non-consented consumption (processing for any reasons) of personal data potentially present in these sources may violate one or more regulations depending on the jurisdiction. 
\paragraph{LLM Outputs}: The second scenario involves the generation of personal data in the output of an LLM application, such as a Gen AI system providing replies to user queries, which contain personal data from one or more people. 
\paragraph{Prompt data in LLM interactions}: Lastly, the third scenario deals with personal data in the construction of a prompt used to interact with an LLM-based Gen AI application. In the latter, the text composing the prompt may willingly or unwillingly contain PII that the LLM may further process to accidentally output more personal data, or even more sensitive data. 

The detection of PII/SPI is critical across these scenarios, and its disposition must align with the enterprise’s Governance, Risk, and Compliance (GRC) directives.

\subsection{Integration of OneShield Guardrails}
To address the risks in these scenarios, our team has deployed the adaptive PII protection within the OneShield Guardrails platform, providing a comprehensive solution for detecting and mitigating PII risks in LLMs. The OneShield guardrails implementation is  deployed internally, and utilizes publicly available unstructured text that is rich in language features. It provides detector analysis across a varied set of detectors, including HAP (Hate, Abuse and Profanity) and copyrighted content, self-harm detectors, etc. This OneShield Guardrails analyzes various risks of model outputs across different models and configure policies. This comprehensive PII approach should help global enterprises to increase their trust in LLMs, at least when it comes to personal data risks. 

\begin{figure}[]
 \centerline{\includegraphics[scale=0.28]{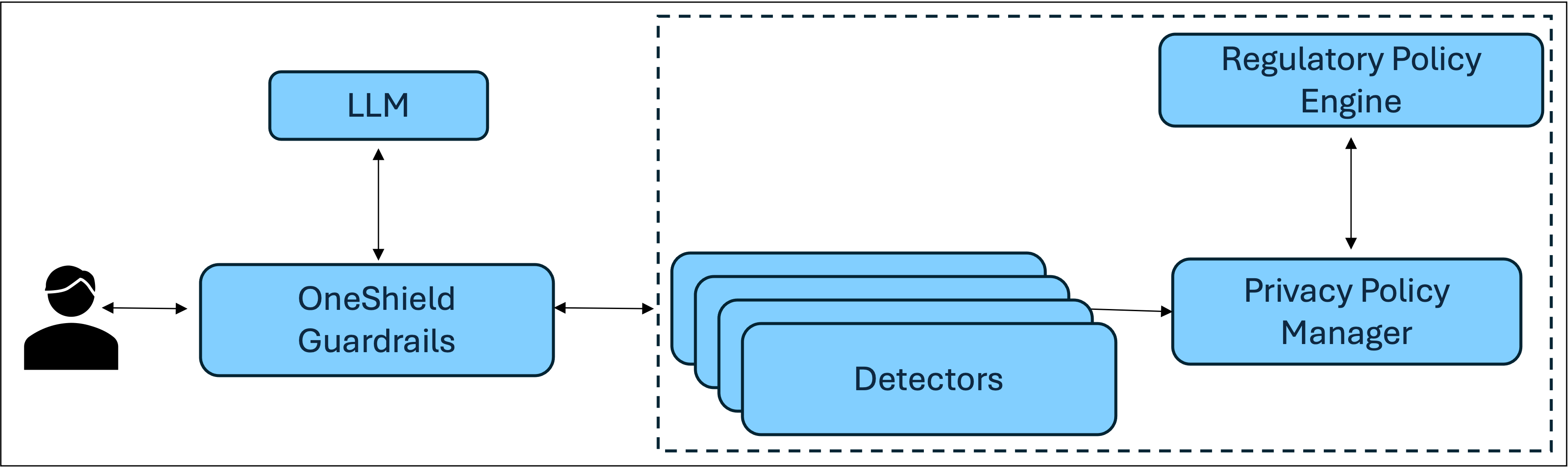}}
\caption{OneShield Guardrails framework}
\label{fig:oneshield}
\end{figure}

The framework has three major functions as seen in Figure \ref{fig:oneshield}: 
\begin{enumerate}
\item \textit{Guardrails:} It covers both input prompt and output, with a flexible, scalable architecture.
\item \textit{Detector Analysis:} Utilises rules-based regulatory policy engine supporting policies including GDPR, CCPA etc.
\item \textit{Policy Manager:} Provides compliance policy templates on detection and potential violations. Violations occur when a detector observes a text/sentence that disobeys the compliance policy.
\end{enumerate}

\begin{figure*}[]
 \centerline{\includegraphics[scale=0.28]{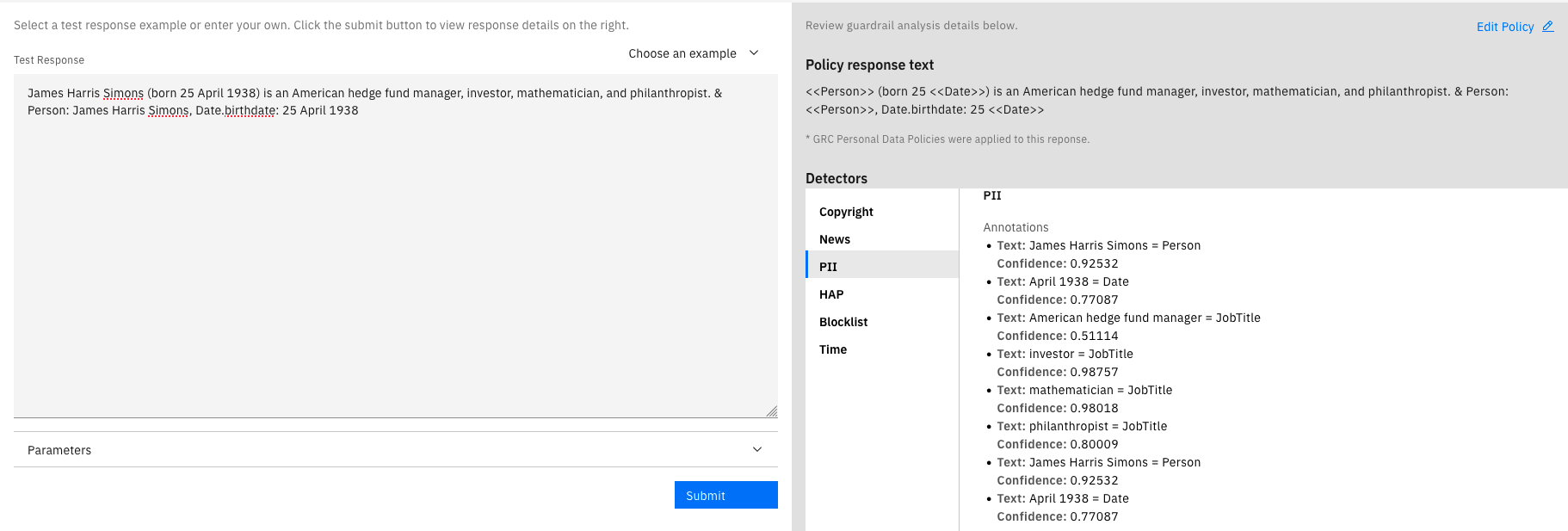}}
\caption{OneShield Guardrails deployment - Adaptive PII system framework}
\label{fig:ui}
\end{figure*}

\paragraph{Deployment workflow} 
This implementation enables global enterprises to build trust in LLMs by proactively addressing personal data risks.
Figure \ref{fig:ui} presents an example of a detected PII entityalong with the adaptive masking approach applied. Based on the rules-based regulatory policy engine, the violations are detected and taken for \textit{Adaptive Masking}. The \textit{Adaptive Masking} masks the PII violations detected. For example, for the test response example in the screenshot: \textit{James Harris Simons (born 25 April 1938) is an American hedge fund manager, investor, mathematician, and philanthropist.}, the \textit{Person, Date, Birth Date} PII was masked.

\section{Validation}
\label{sec:validation}

In this section, we evaluate the performance of the adaptive PII protection framework, a series of benchmarking experiments were conducted. The OneShield PII detector was compared against state-of-the-art tools such as Microsoft Presidio \cite{microsoft} , StarPII \cite{starpii}, and Amazon Comprehend \cite{awscomprehend}.

\subsection{Performance Evaluation}

\begin{table*}[htb]
\renewcommand{\arraystretch}{1.2}
\footnotesize
\centering
\begin{tabular}{p{0.1\textwidth} p{0.1\textwidth} p{0.05\textwidth} p{0.08\textwidth} p{0.1\textwidth} p{0.1\textwidth} p{0.05\textwidth} p{0.08\textwidth} p{0.1\textwidth}}
\hline
 & \multicolumn{4}{c}{\textbf{Benchmark 1 F1}} & \multicolumn{4}{c}{\textbf{Benchmark 2 F1}} \\
 & OneShield PII detector & StarPII & Presidio Analyzer & Comprehend & OneShield PII detector & StarPII & Presidio Analyzer & Comprehend\\
\hline
Person              & \textbf{1}    & 0.99 & 1    & 0.88 & \textbf{0.99} & \textbf{0.99} & 0.87  & 0.87\\
Date                & \textbf{0.94} &      & 0.62 & 0.76 & \textbf{0.98} & & 0.78 & 0.77\\
BankAccount         & 0.73 &      & \textbf{0.79} & 0.66 &      & &     &  \\
CreditCard          & 0.95 &      & \textbf{0.99} & 0.96 &      & &     & \\
EmailAddress        & 0.57 & 1    & \textbf{0.59} & 0.53 & 0.95 &\textbf{1}& 0.95 & 0.96\\
IPAddress           & 0.61 & \textbf{0.7} & 0.47 & 0.56 &      & &     & \\
PassportNumber      & \textbf{0.95} &      & 0.33 & 0.54    & \textbf{1}    & & \textbf{1}   & 0.95\\
PhoneNumber         & \textbf{0.85} &      & 0.1  & 0.74 & 0.65 & & 0.58 & \textbf{0.77}\\
SocialSecurity\\Number& 0.71 &      & 0.06 & \textbf{0.76} &      & &     & \\
Location            & \textbf{0.86} &      & 0.54 & 0.76 & \textbf{0.96} & & 0.67 & 0.93\\
JobTitle            &      &      &      &   & \textbf{0.57} & & 0.18 & \\
\hline
\end{tabular}
\caption{F1 score benchmark results with open source PII detectors}
\label{tab:benchmark}
\end{table*}

\begin{table*}[htb]
\footnotesize
    \centering
\begin{tabular}{p{0.15\textwidth} p{0.15\textwidth} p{0.15\textwidth} p{0.15\textwidth} p{0.2\textwidth}} 
    \toprule
    \textbf{Scenario} & \textbf{Detected PII} & \textbf{Compliance Action (GDPR)} & \textbf{Compliance Action (CCPA)} & \textbf{Outcome}\\
    \midrule
    "David's SSN is 123-45-6789" & SSN & Masked (\#\#\#-\#\#-6789) & Retained but flagged for opt-out & GDPR-Compliant, CCPA-Compliant \\
    "Her passport number is E254" & Passport Number & Fully Anonymized & Fully Anonymized & GDPR-Compliant, CCPA-Compliant \\
    "User John donated \$500" & Name, Donation Amount & Retained (Legitimate Interest) & Masked &GDPR-Compliant, CCPA-Non-Compliant\\
    \bottomrule
\end{tabular}

\caption{Examples of scenarios with detected PII and compliance actions taken under GDPR and CCPA}
\label{tab:scenario}
\end{table*}

\begin{table*}[htb]
\footnotesize
    \centering
\begin{tabular}{p{0.28\textwidth} p{0.15\textwidth} p{0.15\textwidth} p{0.15\textwidth}} 
    \toprule
    \textbf{Sample Sentence} & \textbf{Original} & \textbf{Masked (GDPR)} & \textbf{Masked (CCPA)}\\
    \midrule
    Alice's email is alice\@example.com & alice\@example.com & [email redacted] &[email redacted] \\
    Mark's SSN is 123-45-6789 & 123-45-6789 & \#\#\#-\#\#-6789 & \#\#\#-\#\#-6789\\
    User ID: 456789 & 456789 & XXXXX & XXXXX\\
    \bottomrule
\end{tabular}

\caption{Examples of masking applied to sample sentences under GDPR and CCPA regulations}
\label{tab:sentence}
\end{table*}

\subsubsection{Comparative Benchmarking Against Tools}
 In this section, we provide an overview and results of benchmarking for three datasets on the state-of-the-art PII detectors, namely \cite{microsoft, starpii, awscomprehend}. Microsoft Presidio \cite{microsoft} is an open-source PII detection tool that uses pattern-based and Named Entity Recognition (NER) methods to detect and mask PII in text. StarPII \cite{starpii} is an NER model trained to detect PII in code datasets, taken from HuggingFace. Comprehend \cite{awscomprehend} is Amazon’s NLP service that identifies PII entities in text using machine learning, supporting customizable entity recognition and masking options. 

\textit{Benchmark 1} is a manually annotated in-house dataset with $\sim1500$ test points. \textit{Benchmark 2} is an open-access dataset sourced from a Kaggle challenge \cite{kaggledataset}. Table \ref{tab:benchmark} provides the comparative F1 score values for different PII types for the system OneShield PII detector.  Some of the additional PII types that our work focuses on are not covered by benchmark data, so there are empty spaces. Additionally, not all PII types are covered by the open source detectors, hence there are empty spaces in the table for those PII types. 

The OneShield PII detector demonstrated higher accuracy for nuanced PII types, such as \textit{Passport Numbers, Dates, and Phone Numbers}, where other tools struggled with false positives or low recall rates. For example: Presidio failed to distinguish between \textit{Black Friday} (non-PII) and sensitive dates in specific contexts. Amazon Comprehend missed nested PII like email addresses within hyperlinks \textit{(e.g., http://lore.kernel.org/r/2027-1-jack@suse.cz)}. These results highlight the effectiveness of contextual sensitivity scoring and multilingual adaptation, unique to the OneShield system.

\subsubsection{Observations on Human Evaluation Study for PII and SPI Compliance }

The adaptive PII framework was further evaluated to assess its ability to detect, mask, and anonymize PII/SPI in alignment with the GDPR and the CCPA. This evaluation complements the performance benchmarks with a qualitative and regulatory analysis, addressing gaps highlighted in previous reviews. The study focused on ensuring that the system adheres to compliance requirements while maintaining user trust and operational effectiveness.

\paragraph{Study Design and Objectives} 
The primary objective of this study was to evaluate the system’s detection accuracy, compliance adherence, and user perception of privacy protections. A dataset of 100 annotated examples was curated, representing diverse PII types such as social security numbers (SSNs), names, passport numbers, and credit card information. These examples were drawn from real-world scenarios in both EU and California contexts to test the system's adaptability across regulatory frameworks.

Three key metrics guided the evaluation:
\begin{enumerate}
    \item \textbf{Detection Accuracy}: Measuring the system's ability to correctly identify PII across various contexts and types.
    \item \textbf{Regulatory Compliance}: Assessing whether masking, anonymization, and other remediation techniques align with GDPR and CCPA requirements.
    \item \textbf{User Trust}: Collecting feedback from 20 participants who rated the system's perceived privacy protection on a 5-point scale.
\end{enumerate}


\paragraph{Findings on GDPR and CCPA Compliance} The evaluation revealed notable differences in the system's handling of PII under GDPR and CCPA. GDPR compliance was observed to enforce stricter anonymization for sensitive identifiers like passport numbers. Additionally, the system enabled masking for public figures unless explicitly exempt under legitimate interest clauses. In contrast, CCPA compliance allowed opt-out mechanisms for users to block specific data categories while retaining data utility by applying pseudonymization instead of complete masking, where permissible.

To illustrate these differences, Table \ref{tab:scenario} highlights various scenarios, detected PII, and the compliance actions taken under GDPR and CCPA.

To further demonstrate the system’s performance, Table \ref{tab:sentence} showcases how it applies masking for specific sample sentences under GDPR and CCPA regulations. The adaptive remediation approach ensures that GDPR mandates stricter anonymization for sensitive identifiers, such as replacing email addresses with “[email redacted]” and masking SSNs with hashes (e.g., \#\#\#-\#\#-6789). Meanwhile, CCPA prioritizes pseudonymization and user-driven opt-outs.

\textbf{User Trust Evaluation:} The system's perceived privacy protection capabilities were evaluated by a group of 20 participants. They rated the system on a scale of 1 to 5, resulting in an average score of \textbf{4.6/5}, indicating high user satisfaction. Participants acknowledged the system’s consistent and transparent masking of PII but recommended improved reporting to explain compliance actions clearly.

\textbf{Enhanced Interface for Remediation Actions:} The user interface for the OneShield adaptive remediation system further demonstrated its effectiveness. As illustrated in Figure \ref{fig:ui}, detected violations are dynamically masked according to compliance rules. This interface provides users with visibility into detected PII and the applied remediation actions, reinforcing trust and transparency.

Summarising, the human evaluation study highlighted the adaptive PII framework’s ability to navigate the nuanced requirements of GDPR and CCPA while maintaining user trust. By balancing strict anonymization mandates with data utility, the system enables enterprises to achieve compliance without compromising operational goals. These findings validate the system’s potential as a robust and scalable solution for mitigating privacy risks in diverse regulatory environments.

\section{Conclusions and Future Work}
\label{sec:conclusion}
In this paper, we present an adaptive system designed to mitigate the risks associated with PII in LLMS while ensuring compliance with global data protection regulations. By leveraging NLP techniques, real-time contextual analysis, and a policy-driven masking framework, the system dynamically adapts to the regulatory requirements of different jurisdictions. This adaptability empowers enterprises to deploy LLMs confidently, mitigating risks of PII exposure while adhering to governance, risk, and compliance (GRC) policies.

The system’s versatility is demonstrated through its applicability across diverse sectors, such as healthcare and finance, where regulatory demands are stringent and sector-specific. By reducing the likelihood of unintentional PII exposure, the framework promotes ethical and compliant AI deployment, fostering trust in LLM-powered applications.

While the system demonstrates significant potential, it must address challenges such as the continuous evolution of data protection laws and the complexity of integrating conflicting regulations across jurisdictions. For instance, handling ambiguous data—like distinguishing public figures' names in news articles from sensitive personal references—requires improved edge case handling.

We plan to enhance the system by refining compliance mechanisms to reconcile overlapping regulatory requirements, such as GDPR’s stringent anonymization versus CCPA’s opt-out provisions. Expansion into additional domains, including telecommunications, retail, and government, will involve collaboration with domain experts to fine-tune detection and remediation rules.

Efforts are underway to automate the integration of new regulations into the policy engine and improve the system’s efficiency for processing high-volume, real-time LLM interactions. These enhancements will ensure the system remains a robust, scalable solution for global PII protection, enabling ethical and compliant LLM deployment.

\section{Acknowledgements}
\label{sec:ack}
We would like to thank Guang-Jie Ren, Pawan Chowdhary, and Sandeep Gopisetty for their guidance on this work.


\bibliography{aaai25}

\end{document}